\documentclass{article}
\PassOptionsToPackage{numbers, compress}{natbib}
\usepackage[preprint]{neurips_2024}

\usepackage[utf8]{inputenc} 
\usepackage[T1]{fontenc}    
\usepackage{hyperref}       
\usepackage{url}            
\usepackage{booktabs}       
\usepackage{amsfonts}       
\usepackage{nicefrac}       
\usepackage{microtype}      
\usepackage{xcolor}         

\usepackage{times}
\usepackage{CJKutf8}
\usepackage{latexsym}
\usepackage{tcolorbox}
\usepackage{multirow}
\usepackage{wrapfig}
\usepackage{tikz}
\usepackage{capt-of}
\usepackage{graphicx}  
\usepackage{pgfplots}
\pgfplotsset{compat=1.12}
\usepackage{amsmath}
\usepackage{multicol}
\usepackage{color}
\usepackage{mwe}
\usepackage{wrapfig}
\usepackage{colortbl,array}
\usepackage{xspace}
\usepackage{tikz}
\usetikzlibrary{tikzmark}
\makeatletter
\newcommand*\myfontsize{%
  \@setfontsize\myfontsize{7}{8}%
}
\makeatother
\newcommand{\mytextbox}[2]{\tikzmarknode[draw=#1,thick,inner sep=2pt]{test}{\myfontsize #2}}
\definecolor{myred}{rgb}{0.7, 0.3, 0.0}
\definecolor{myblue}{HTML}{054488}
\definecolor{mygreen}{HTML}{056b34}

\newcommand{\blue}[1]{\mytextbox{myblue}{\textbf{\textcolor{myblue}{#1}}}}
\newcommand{\green}[1]{\mytextbox{mygreen}{\textbf{\textcolor{mygreen}{#1}}}}

\newcolumntype{R}[1]{>{\raggedleft\let\newline\\\arraybackslash\hspace{0pt}}m{#1}}

\usetikzlibrary{intersections}

\definecolor{darkgreen}{rgb}{0.0, 0.42, 0.24}
\usepackage{caption}
\usepackage{subcaption}
\usepackage{graphicx}
\usepackage{pifont}
\usepackage{titletoc}
\usepackage{amsfonts}
\usepackage{booktabs}
\usepackage{arydshln}
\usepackage{colortbl}
\usepackage{algorithm}
\usepackage[noend]{algpseudocode}
\usepackage{enumitem}
\usepackage{graphicx}
\usepackage{soul}
\usepackage{colortbl,array,xcolor}

\definecolor{citecolor}{HTML}{0051f4}
\definecolor{pink}{HTML}{ed008c}
\hypersetup{
    colorlinks,
    linkcolor=citecolor,
    urlcolor=pink,
    citecolor=citecolor
}

\title{Search-o1: Agentic Search-Enhanced\\ Large Reasoning Models}

\author{
Xiaoxi Li$^1$, Guanting Dong$^1$, Jiajie Jin$^1$, Yuyao Zhang$^1$, Yujia Zhou$^2$, \\
\textbf{Yutao Zhu$^1$, Peitian Zhang$^1$, Zhicheng Dou$^1$\thanks{Correpsonding author.}} \\
$^1$Renmin University of China ~~ $^2$Tsinghua University\\
\texttt{\{xiaoxi\_li, dou\}@ruc.edu.cn} \\
\\
Project Page: \url{https://search-o1.github.io/}
}

\begin{document}
\begin{CJK}{UTF8}{gbsn}

\maketitle

\begin{abstract}

Large reasoning models (LRMs) like OpenAI-o1 have demonstrated impressive long stepwise reasoning capabilities through large-scale reinforcement learning. However, their extended reasoning processes often suffer from knowledge insufficiency, leading to frequent uncertainties and potential errors. To address this limitation, we introduce \textbf{Search-o1}, a framework that enhances LRMs with an agentic retrieval-augmented generation (RAG) mechanism and a Reason-in-Documents module for refining retrieved documents. Search-o1 integrates an agentic search workflow into the reasoning process, enabling dynamic retrieval of external knowledge when LRMs encounter uncertain knowledge points. Additionally, due to the verbose nature of retrieved documents, we design a separate Reason-in-Documents module to deeply analyze the retrieved information before injecting it into the reasoning chain, minimizing noise and preserving coherent reasoning flow. Extensive experiments on complex reasoning tasks in science, mathematics, and coding, as well as six open-domain QA benchmarks, demonstrate the strong performance of Search-o1. This approach enhances the trustworthiness and applicability of LRMs in complex reasoning tasks, paving the way for more reliable and versatile intelligent systems.
The code is available at \url{https://github.com/sunnynexus/Search-o1}.

\end{abstract}
\section{Introduction}

Recently emerged large reasoning models (LRMs), exemplified by OpenAI's o1~\cite{openai2024openaio1card}, Qwen-QwQ~\cite{qwq-32b-preview} and DeepSeek-R1~\cite{deepseek-r1}, employ large-scale reinforcement learning foster impressive long-sequence stepwise reasoning capabilities, offering promising solutions to complex reasoning problems~\cite{openai2024reasoning,reason1,wei2022chain,zhong2024evaluation,metamath,rft,qwen2.5math}. This advancement has inspired a series of foundational efforts aimed at exploring and reproducing o1-like reasoning patterns, to broaden their application to a wider range of foundational models~\cite{2410_o1_journey_part1,huang2024o1,zhang2024llamaberry,zhang2024o1coder,ye2024physics,jiang2024technical,2412_min_imitate}.

It is noteworthy that o1-like reasoning patterns guide LRMs to engage in a slower thinking process~\cite{daniel2017thinking,wu2024comparative} by implicitly breaking down complex problems, generating a long internal reasoning chain and then discovering suitable solutions step by step.  While this characteristic enhances logical coherence and interpretability of reasoning, an extended chain of thought may cause overthinking~\cite{chen2024think23overthinkingo1like} and increased risks of knowledge insufficiency~\cite{uncertain1,uncetain2,uncertain3}, where any knowledge gap can propagate errors and disrupt the entire reasoning chain~\cite{DBLP:journals/corr/abs-2309-01219,error1,error2,reason_rag}.

\begin{figure}[!t]
\centering
\begin{subfigure}[htbp]{0.485\textwidth}
\centering
\fontsize{9.3pt}{10.5pt}\selectfont
\vspace{-12pt}
\begin{tabular}{p{0.9\textwidth}}
    \toprule
    \textbf{Cases of Model-Expressed Uncertainty} \\
    \midrule
    \textbf{Wait,} \textbf{perhaps} it's referring to dimethyl sulfone, but that doesn't seem right. \\
    \midrule
    \textbf{Alternatively}, \textbf{perhaps} there's a mistake in my understanding of epistasis. Let me look up epistasis quickly. Epistasis is ... \\
    \midrule
    \textbf{Alternatively}, HBr could also abstract a hydrogen atom from the alkene, leading to a ... \\
    \midrule
    As I recall, Quinuclidine is a seven-membered ring with a nitrogen atom, \textbf{likely} not having the required symmetry. \\
    \bottomrule
\end{tabular}
\label{tab:uncertain_words}
\end{subfigure}
\hfill
\begin{subfigure}[htbp]{0.485\textwidth}
    \centering
    \includegraphics[width=\linewidth]{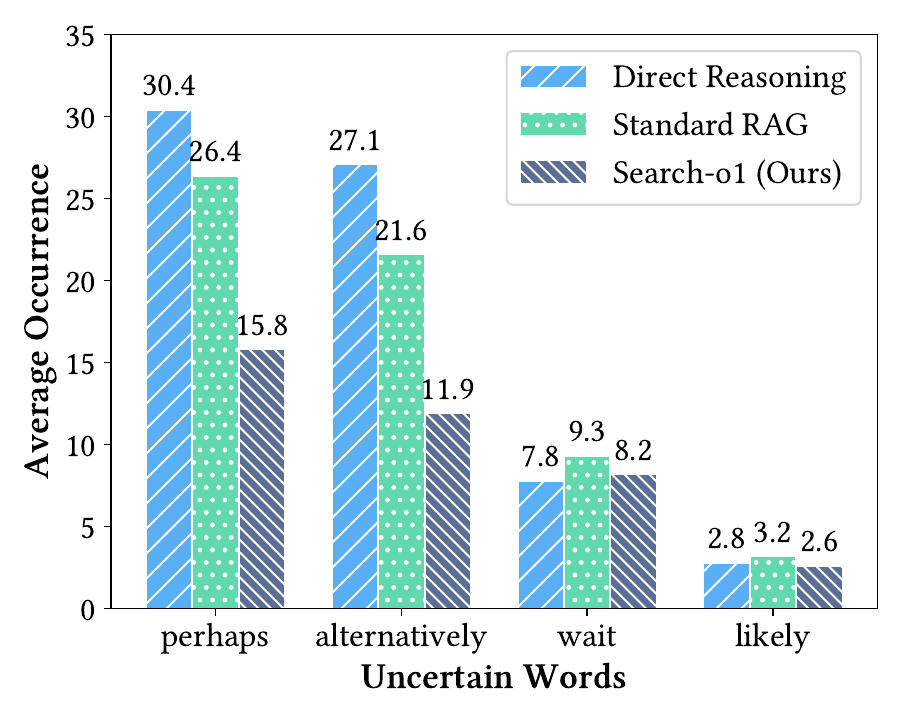}
    \label{fig:words_statistics}
\end{subfigure}
\vspace{-7.5pt}
\caption{
Analysis of reasoning uncertainty with QwQ-32B-Preview. \textbf{Left}: Examples of uncertain words identified during the reasoning process. \textbf{Right}: Average occurrence of high-frequency uncertain words per output in the GPQA diamond set.
}
\label{fig:combined}
\end{figure}

To address this limitation, we conduct preliminary experiments to assess the frequency of uncertain words decoded by the LRMs due to knowledge gaps. As shown in Figure~\ref{fig:combined}, the extended thinking process leads LRM to frequently decode numerous uncertain terms in challenging reasoning problems, with ``\textit{perhaps}'' averaging over 30 occurrences in each reasoning process. Notably, the high specialization of these problems also complicates manual reasoning verification, often incurring significant costs~\cite{human1}. Consequently, automating the supplementation of knowledge required for the o1-like reasoning process has become a significant challenge, limiting the progress of LRMs in achieving universally trustworthy reasoning.

To shed light on this topic, our core motivation is to enhance the LRMs with o1-like reasoning pattern through autonomous retrieval. We propose \textbf{Search-o1}, which integrates the reasoning process of LRMs with two core components: an agentic retrieval-augmented generation (RAG) mechanism and a knowledge refinement module. This design aims to enable LRMs to incorporate the agentic search workflow into the reasoning process, retrieving external knowledge on demand to support step-wise reasoning while preserving coherence throughout. 

Specifically, our results in Figure~\ref{fig:combined} reveal that traditional problem-oriented RAG techniques do not effectively address the knowledge gaps compared to direct reasoning (Standard RAG vs. Direct Reasoning). This finding aligns with human intuition, as standard RAG retrieves relevant knowledge only once in a problem-oriented manner, while the knowledge required for each step in complex reasoning scenarios is often varied and diverse~\cite{zheng2024processbench,reason_rag,2412_AR_MCTS}. Unlike them, Search-o1 employs an agentic RAG technique that guides the model to actively decode search queries when facing knowledge shortages, thereby triggering the retrieval mechanism to obtain relevant external knowledge. Owing to the benefits of this design, our retrieval mechanism can be triggered and iterated multiple times within a single reasoning session to fulfill the knowledge needs of various reasoning steps.

To effectively integrate retrieved knowledge into the LRM's reasoning process, we further identify two key challenges when directly incorporating retrieved documents into the reasoning chain during practical experiments: 
(1) \textbf{Redundant Information in Retrieved Documents.} Retrieved documents are often lengthy and contain redundant information, directly inputting them into LRMs may disrupt the original coherence of reasoning and even introduce noise~\cite{wu2024easilyirrelevantinputsskew, YoranWRB24, longcontext1}.
(2) \textbf{Limited Ability to Understand Long Documents.} Most LRMs have been specifically aligned for complex reasoning tasks during the pre-training and fine-tuning stages. This focus has resulted in a degree of catastrophic forgetting in their general capabilities~\cite{forgot1,how_abilities}, ultimately limiting their long-context understanding of retrieved documents.


To address these challenges, we introduce the Reason-in-Documents module, which operates independently from the main reasoning chain. This module first conducts a thorough analysis of retrieved documents based on both the current search query and previous reasoning steps, and then produces refined information that seamlessly integrates with the prior reasoning chain.


In summary, our contributions are as follows:

\begin{itemize}[leftmargin=1em]
\item We propose Search-o1, the first framework that integrates the agentic search workflow into the o1-like reasoning process of LRM for achieving autonomous knowledge supplementation.

\item To effectively integrate external knowledge during reasoning, Search-o1 combines the reasoning process with an agentic RAG mechanism and a knowledge refinement module. This design enables the LRM to retrieve external knowledge on demand, seamlessly incorporating it into the reasoning chain while maintaining the original logical flow.

\item With five complex reasoning domains and six open-domain QA benchmarks, we demonstrate that Search-o1 achieves remarkable performance in the reasoning field while maintaining substantial improvements in the general knowledge. Further quantitative analysis confirms its efficiency and scalability, offering practical guidance for trustworthy reasoning in LRMs.
\end{itemize}

\section{Related Work}

\paragraph{Large Reasoning Models.} 
Large reasoning models focus on enhancing performance at test time by utilizing extended reasoning steps, contrasting with traditional large pre-trained models that achieve scalability during training by increasing model size or expanding training data~\cite{henighan2020scaling,qwen2,qwen2.5,2409_o1_eval, 2412_o1_roadmap}. Studies have shown that test-time scaling can improve the reasoning abilities of smaller models on complex tasks~\cite{feng2024towards, zelikman2024quiet}. Recently, models like OpenAI-o1~\cite{openai2024openaio1card}, Qwen-QwQ~\cite{qwq-32b-preview} and DeepSeek-R1~\cite{deepseek-r1} explicitly demonstrate chain-of-thought reasoning~\cite{wei2022chain}, mimicking human problem-solving approaches in domains such as mathematics, coding, and so on.

Various approaches have been explored to achieve o1-like reasoning capabilities. Some methods combine policy and reward models with Monte Carlo Tree Search (MCTS) \cite{jiang2024technical}, though this does not internalize reasoning within the model. Other studies incorporate deliberate errors in reasoning paths during training to partially internalize these abilities \cite{2410_o1_journey_part1, ye2024physics}. Additionally, distilling training data has been shown to enhance models' o1-like reasoning skills \cite{2412_min_imitate}. The o1-like reasoning paradigm has demonstrated strong performance across diverse domains, including vision-language reasoning \cite{2411_llava_o1, 2412_AR_MCTS, wemath, 2412_Mulberry}, code generation \cite{2412_o1_coder,dmath}, healthcare \cite{2412_HuatuoGPT_o1}, and machine translation \cite{2412_DRT_o1}. However, these approaches are limited by their reliance on static, parameterized models, which cannot leverage external world knowledge when internal knowledge is insufficient.

\paragraph{Retrieval-Augmented Generation.}
Retrieval-augmented generation (RAG) introduces retrieval mechanisms to address the limitations of static parameters in generative models, allowing access to external knowledge to solve more complex problems~\cite{lewis2020retrieval, zhao2024retrieval, genir_survey, 2409_trustworthy_rag_survey}. Advanced research in this field enhances the RAG system from multiple aspects, including the necessity of retrieval~\cite{tan2024small}, pre-processing of queries~\cite{ma2023query, wang2023query2doc}, retrieved documents compressing~\cite{xu2023recomp}, denoising~\cite{2410_can_rag_help_reason,dparag}, refining~\cite{jiang2023longllmlingua, jin2024bider, MetaRAG}, instruction following~\cite{followrag,autoif, 2411_AssistRAG} and so on. Furthermore, some studies have explored end-to-end model training to implement RAG systems~\cite{selfrag, unigen, corpuslm, retrollm} and knowledge-graph-based RAG systems~\cite{edge2024localglobalgraphrag, liang2024kagboostingllmsprofessional}.

Recently, agentic RAG systems empower models to autonomously determine when and what knowledge to retrieve as needed, showcasing enhanced planning and problem-solving capabilities~\cite{chen2024mindsearch, verma2024plan, yao2022react}. There is also research combining agent-based systems with MCTS to optimize complex workflows, leveraging retrievers and other tools to accomplish tasks~\cite{zhang2024aflow}. However, existing RAG approaches have not combined the strong reasoning capabilities of o1-like models, limiting the potential to further enhance system performance in solving complex tasks.

\begin{figure*}[!t]
\centering
\includegraphics[width=0.985\linewidth]{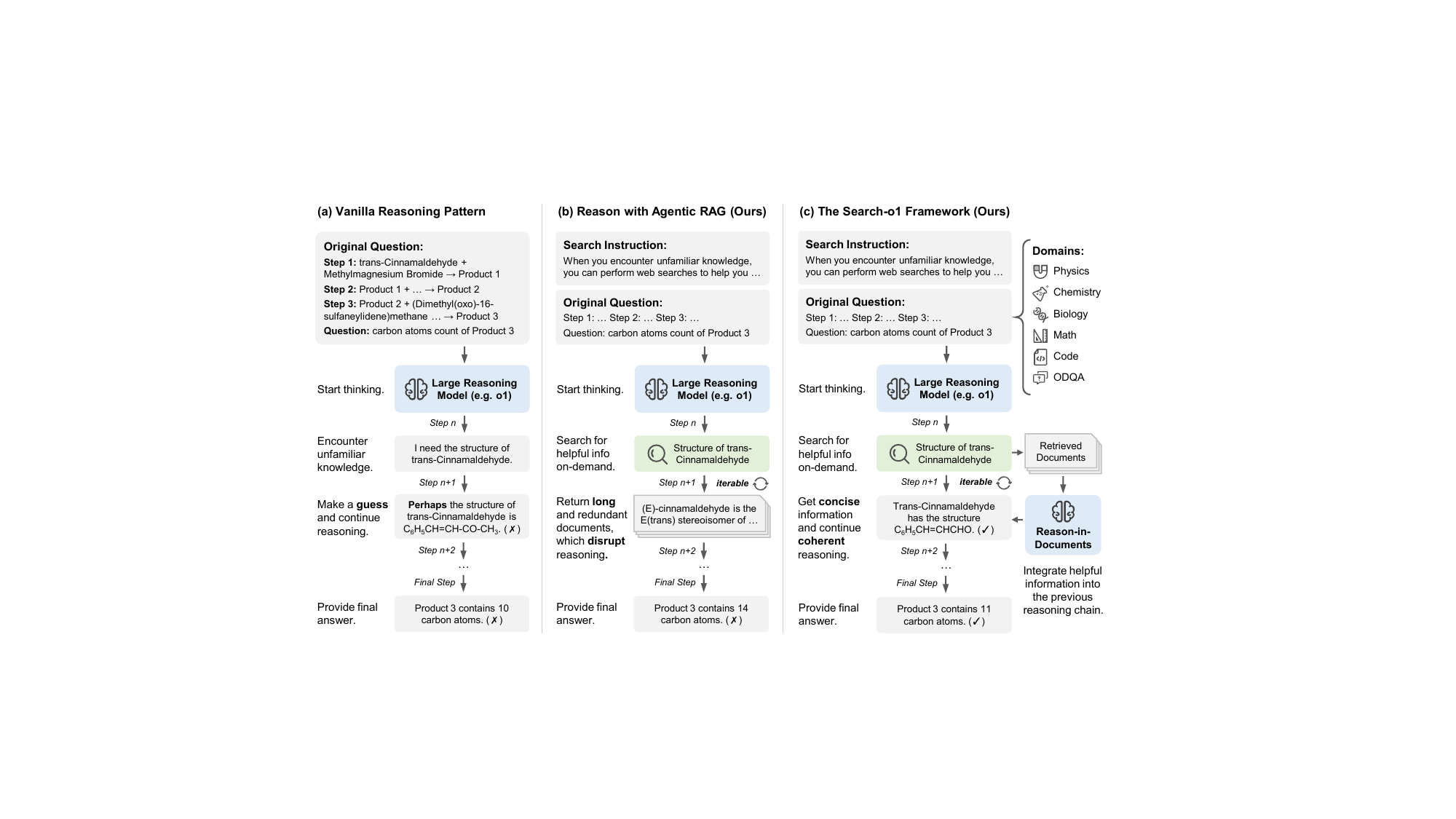}
\caption{
Comparison of reasoning approaches: (a) Direct reasoning without retrieval often results in inaccuracies due to missing knowledge. (b) Our agentic retrieval-augmented reasoning approach improves knowledge access but usually returns lengthy, redundant documents, disrupting coherent reasoning. (c) Our Search-o1 integrates concise and accurate retrieved knowledge seamlessly into the reasoning process, enabling precise and coherent problem-solving.
}
\label{fig:compare} 
\end{figure*}

\section{Methodology}

\subsection{Problem Formulation}
We consider a complex reasoning task that necessitates multi-step reasoning and the retrieval of external knowledge to derive solutions. The objective is to generate a comprehensive solution for each question $q$, consisting of both a logical reasoning chain $\mathcal{R}$ and the final answer $a$. In this work, we enable the reasoning model to utilize external knowledge sources during the reasoning process. Specifically, we consider three primary inputs in the problem-solving process: the task instruction $I$, the question $q$, and externally retrieved documents $\mathcal{D}$. Here, $I$ provides an overarching description of the reasoning task, $q$ is the specific complex question to be answered, and $\mathcal{D}$ comprises background knowledge dynamically retrieved from a relevant knowledge base. 

The goal is to design a reasoning mechanism that effectively integrates $I$, $q$, and $\mathcal{D}$ to produce a coherent reasoning chain $\mathcal{R}$ and a final answer $a$. This can be formalized as the mapping $(I, q, \mathcal{D}) \rightarrow (\mathcal{R}, a)$. The generation of the reasoning sequence and the final answer can be expressed as:
\begin{equation}
P(\mathcal{R}, a \mid I, q, \mathcal{D}) = \underbrace{\prod_{t=1}^{T_r} P(\mathcal{R}_{t} \mid \mathcal{R}_{<t}, I, q, \mathcal{D}_{<t})}_{\text{Reasoning Process}} \cdot \underbrace{\prod_{t=1}^{T_a} P(a_t \mid a_{<t}, \mathcal{R}, I, q)}_{\text{Answer Generation}},
\label{equ:origin_reason_ans}
\end{equation}
where $T_r$ is the number of tokens in the reasoning sequence $\mathcal{R}$. The token at the position $t$ is $\mathcal{R}_t$, and $\mathcal{R}_{<t}$ represents all tokens generated before position $t$. \( \mathcal{D}_{\leq t} \) represents all documents retrieved up to token \( t \) in the reasoning chain. Similarly, $T_a$ is the length of the answer sequence $a$, with $a_t$ being the token at the position $t$ and $a_{<t}$ indicating all generated answer tokens before the position $t$.

\subsection{Overview of the Search-o1 Framework}

The Search-o1 framework addresses knowledge insufficiency in large reasoning models (LRMs) by seamlessly integrating external knowledge retrieval into their reasoning process while maintaining chain-of-thought coherence. As illustrated in Figure~\ref{fig:compare}, we present a comparative analysis of three approaches: vanilla reasoning, agentic retrieval-augmented generation (RAG), and our proposed Search-o1 framework.

\begin{itemize}[leftmargin=1em]
\item \textbf{Vanilla Reasoning Pattern:} Consider the example in Figure~\ref{fig:compare}(a), where the task involves determining the carbon atom count in the final product of a three-step chemical reaction. The vanilla reasoning approach falters when encountering knowledge gaps (e.g., the ``structure of trans-Cinnamaldehyde''). Without access to accurate information, the model must rely on assumptions, potentially leading to cascading errors throughout subsequent reasoning steps.

\item \textbf{Agentic RAG:} To bridge the knowledge gaps during reasoning, we build the agentic RAG mechanism (Figure~\ref{fig:compare}(b)) to enable the model to autonomously retrieve external knowledge when needed. When uncertainty arises—such as regarding the compound's structure—the model generates targeted search queries (e.g., ``structure of trans-Cinnamaldehyde''). However, the direct insertion of retrieved documents, which often contain lengthy and irrelevant information, may disrupt the reasoning flow and hurt coherence.

\item \textbf{Search-o1:} Our Search-o1 framework (Figure~\ref{fig:compare}(c)) extends the agentic RAG mechanism by incorporating a Reason-in-Documents module. This module condenses retrieved documents into focused reasoning steps that integrate external knowledge while maintaining the logical flow of the reasoning chain. It considers the current search query, retrieved documents, and the existing reasoning chain to generate coherent steps. This iterative process continues until the final answer is reached. The following sections provide detailed explanations of agentic RAG, Reason-in-Documents, and the Search-o1 inference process.

\end{itemize}

\subsection{Agentic Retrieval-Augmented Generation Mechanism}
\label{sec:rag_agent}

The agentic RAG mechanism is a pivotal component of the Search-o1 framework, empowering the reasoning model to autonomously determine when to retrieve external knowledge during the reasoning process. This mechanism allows the model itself to decide whether to continue generating reasoning steps or to initiate a retrieval step. Detailed model instructions can be found in Appendix~\ref{app:instruction_search_o1}.

During the generation of the reasoning chain $\mathcal{R}$, the model may intermittently generate search queries $q_{\text{search}}^{(i)}$ encapsulated between special symbols \green{{<|begin\_search\_query|>}} and \green{{<|end\_search\_query|>}}, where $i$ indexes the $i$-th search step. Each search query is generated based on the current state of the reasoning process and the previously retrieved knowledge. The generation of each search query is expressed as:
\begin{equation}
P(q_{\text{search}}^{(i)} \mid I, q, \mathcal{R}^{(i-1)}) = \prod_{t=1}^{T_q^{(i)}} P\left(q_{\text{search}, t}^{(i)} \mid q_{\text{search}, <t}^{(i)}, I, q, \mathcal{R}^{(i-1)}\right),
\end{equation}
where $T_q^{(i)}$ is the length of the $i$-th search query, $q_{\text{search}, t}^{(i)}$ denotes the token generated at step $t$ of the $i$-th search query, and $\mathcal{R}^{(i-1)}$ represents all the reasoning steps prior to the $i$-th search step, including both search queries and search results.

Once a new pair of special symbols for the search query is detected in the reasoning sequence, we pause the reasoning process, and the search query $q_{\text{search}}^{(i)}$ is extracted. The retrieval function $\texttt{Search}$ is invoked to obtain relevant documents:
\begin{equation}
\mathcal{D}^{(i)} = \texttt{Search}(q_{\text{search}}^{(i)}),
\end{equation}
where $\mathcal{D}^{(i)} = {d_1^{(i)}, d_2^{(i)}, \ldots, d_{k_i}^{(i)}}$ represents the set of top-$k_i$ relevant documents retrieved for the $i$-th search query.
The retrieved documents $\mathcal{D}^{(i)}$ are subsequently injected into the reasoning chain $\mathcal{R}^{(i-1)}$ between the special symbols \blue{{<|begin\_search\_result|>}} and \blue{{<|end\_search\_result|>}}, allowing the reasoning model to utilize the external knowledge to continue the reasoning process. 

This agentic mechanism enables the model to dynamically and efficiently incorporate external knowledge, maintaining the coherence and relevance of the reasoning process while avoiding information overload from excessive or irrelevant retrieval results.

\subsection{Knowledge Refinement via Reason-in-Documents}

While the agentic RAG mechanism addresses knowledge gaps in reasoning, directly inserting full documents can disrupt coherence due to their length and redundancy. To overcome this, the Search-o1 framework includes the knowledge refinement module, which selectively integrates only relevant and concise information into the reasoning chain through a separate generation process using the original reasoning model. This module processes retrieved documents to align with the model's specific reasoning needs, transforming raw information into refined, pertinent knowledge while maintaining coherence and logical consistency of the main reasoning chain.

The refinement guidelines for Reason-in-Documents are detailed in Appendix~\ref{app:instruction_search_o1}. These guidelines instruct the model to analyze the retrieved web pages based on the previous reasoning steps, current search query, and the content of the searched web pages. The objective is to extract relevant and accurate information that directly contributes to advancing the reasoning process for the original question, ensuring seamless integration into the existing reasoning chain.

For each search step $i$, let $\mathcal{R}^{(<i)}$ denote the reasoning chain accumulated up to just before the $i$-th search query. Given $\mathcal{R}^{(<i)}$, the current search query $q_{\text{search}}^{(i)}$, and the retrieved documents $\mathcal{D}^{(i)}$, the knowledge refinement process operates in two stages: first generating an intermediate reasoning sequence ${r}_{\text{docs}}^{(i)}$ to analyze the retrieved documents, then producing refined knowledge ${r}_{\text{final}}^{(i)}$ based on this analysis. The generation of the intermediate reasoning sequence ${r}_{\text{docs}}^{(i)}$ is expressed as:
\begin{equation}
P({r}_{\text{docs}}^{(i)} \mid \mathcal{R}^{(<i)}, q_{\text{search}}^{(i)}, \mathcal{D}^{(i)}) = \prod_{t=1}^{T_d^{(i)}} P\left({r}_{\text{docs}, t}^{(i)} \mid {r}_{\text{docs}, <t}^{(i)}, \mathcal{R}^{(<i)}, q_{\text{search}}^{(i)}, \mathcal{D}^{(i)}\right),
\end{equation}
where $T_d^{(i)}$ is the length of the intermediate reasoning sequence, and ${r}_{\text{docs}, t}^{(i)}$ denotes the token at step $t$. The refined knowledge ${r}_{\text{final}}^{(i)}$ is then generated based on this analysis:
\begin{equation}
P({r}_{\text{final}}^{(i)} \mid {r}_{\text{docs}}^{(i)}, \mathcal{R}^{(<i)}, q_{\text{search}}^{(i)}) = \prod_{t=1}^{T_r^{(i)}} P\left({r}_{\text{final}, t}^{(i)} \mid {r}_{\text{final}, <t}^{(i)}, {r}_{\text{docs}}^{(i)}, \mathcal{R}^{(<i)}, q_{\text{search}}^{(i)}\right),
\end{equation}
where $T_r^{(i)}$ is the length of the refined knowledge sequence, and ${r}_{\text{final}, t}^{(i)}$ denotes the token at step $t$. The refined knowledge ${r}_{\text{final}}^{(i)}$ is then incorporated into the reasoning chain $\mathcal{R}^{(i)}$, enabling the model to continue generating coherent reasoning steps with access to the external knowledge.
\begin{equation}
P(\mathcal{R}, a \mid I, q) = \prod_{t=1}^{T_r} P\left(\mathcal{R}_{t} \mid \mathcal{R}_{<t}, I, q, \{r_{\text{final}}^{(j)}\}_{j\leq i(t)}\right) \cdot \prod_{t=1}^{T_a} P\left(a_t \mid a_{<t}, \mathcal{R}, I, q\right),
\end{equation}
where $\{r_{\text{final}}^{(j)}\}_{j\leq i(t)}$ denotes all previously refined knowledge up to the $i(t)$-th search step. Here, $i(t)$ represents the index of the search step corresponding to the current reasoning step $t$. This refined knowledge integration ensures that each reasoning step can access relevant external information while maintaining the conciseness and focus of the reasoning process.

\begin{algorithm}[t]
\fontsize{9.6pt}{10.8pt}\selectfont
\caption{Search-o1 Inference}\label{algo:searcho1_inference}
\begin{algorithmic}[1]
\Require Reasoning Model $\mathcal{M}$, Search function \texttt{Search}
\State \textbf{Input:} Questions $\mathcal{Q}$, Task instruction $I$, Reason-in-documents instruction $I_\text{docs}$
\State Initialize set of unfinished sequences $\mathcal{S} \gets \{ I \oplus q \mid q \in \mathcal{Q} \}$
\State Initialize set of finished sequences $\mathcal{F} \gets \{\}$
\While{$\mathcal{S} \neq \emptyset$}
    \State Generate all sequences in $\mathcal{S}$ until EOS or \green{<|end\_search\_query|>}: $\mathcal{T} \gets \mathcal{M}(\mathcal{S})$  \Comment{Batch Generate}
    \State Initialize empty set $\mathcal{S}_r \gets \{\}$ \Comment{Reason-in-documents Inputs}
    \For{each sequence $\text{Seq} \in \mathcal{T}$}
        \If{$\text{Seq}$ ends with \green{<|end\_search\_query|>}}
            \State Extract search query: $q_\text{search} \gets \text{Extract}(\text{Seq}, \green{<|begin\_search\_query|>}, \green{<|end\_search\_query|>})$
            \State Retrieve documents: $\mathcal{D} \gets \texttt{Search}(q_\text{search})$ \Comment{Retrieval}
            \State Construct input for Reason-in-documents: $I_\mathcal{D} \gets I_{\text{docs}} \oplus q_\text{search} \oplus \text{Seq}$
            \State \textbf{Append} the tuple $(I_\mathcal{D}, \text{Seq})$ to $\mathcal{S}_r$
        \ElsIf{$\text{Seq}$ ends with EOS}
            \State Remove $\text{Seq}$ from $\mathcal{S}$, add $\text{Seq}$ to $\mathcal{F}$ \Comment{Sequence Finished}
        \EndIf
    \EndFor
    \If{$\mathcal{S}_r \neq \emptyset$}
        \State Prepare batch inputs: $\mathcal{I}_r \gets \{ I_\mathcal{D} \mid (I_\mathcal{D}, \text{Seq}) \in \mathcal{S}_r \}$
        \State Reason-in-documents: $\mathcal{T}_r \gets \mathcal{M}(\mathcal{I}_r)$ \Comment{Batch Generate}
        \For{$i \gets \{1, ..., |\mathcal{T}_r|\}$}
            \State Let $r \gets \mathcal{T}_r[i]$, $\text{Seq} \gets \mathcal{S}_r[i].\text{Seq}$
            \State Extract knowledge-injected reasoning step: $r_\text{final} \gets \text{Extract}(r)$
            \State Update sequence in $\mathcal{S}$: $\text{Seq} \gets \text{Insert}\left(\blue{<|begin\_search\_result|>}, r_\text{final}, \blue{<|end\_search\_result|>}\right)$
        \EndFor
    \EndIf
\EndWhile
\State \textbf{Output:} Finished Sequences $\mathcal{F}$
\end{algorithmic}
\end{algorithm}

\subsection{Search-o1 Inference Process}

\paragraph{Inference Logic for a Single Question.}
For each question, the Search-o1 inference begins by initializing the reasoning sequence with the task instruction $I$ concatenated with the specific question $q$. As the reasoning model $\mathcal{M}$ generates the reasoning chain $\mathcal{R}$, it may produce search queries encapsulated within the special symbols \green{{<|begin\_search\_query|>}} and \green{{<|end\_search\_query|>}}. Upon detecting the \green{{<|end\_search\_query|>}} symbol, the corresponding search query $q_{\text{search}}$ is extracted, triggering the retrieval function \texttt{Search} to obtain relevant external documents $\mathcal{D}$. These retrieved documents, along with the reason-in-documents instruction $I_{\text{docs}}$ and the current reasoning sequence $\mathcal{R}$, are then processed by the Reason-in-Documents module. This module refines the raw documents into concise, pertinent information $r_{\text{final}}$, which is seamlessly integrated back into the reasoning chain $\mathcal{R}$ within symbols \blue{{<|begin\_search\_result|>}} and \blue{{<|end\_search\_result|>}}. This iterative process ensures that the reasoning model incorporates necessary external knowledge while maintaining coherence and logical consistency, leading to the generation of a comprehensive reasoning chain $\mathcal{R}$ and the final answer $a$.

\paragraph{Batch Inference Mechanism.}
To efficiently handle multiple questions simultaneously, the Search-o1 framework employs a batch inference mechanism that optimizes both token generation and knowledge refinement. Initially, a set of unfinished reasoning sequences $\mathcal{S}$ is created by concatenating the task instruction $I$ with each question $q$ in the batch $\mathcal{Q}$. The reasoning model $\mathcal{M}$ then generates tokens for all sequences in $\mathcal{S}$ in parallel, advancing each reasoning chain until it either completes or requires external knowledge retrieval. When a search query is identified within any sequence, the corresponding queries are extracted and processed in batches through the \texttt{Search} function to retrieve relevant documents $\mathcal{D}$. These documents are then collectively refined by the Reason-in-Documents module, which generates the refined knowledge $r_{\text{final}}$ for each sequence. The refined knowledge is subsequently inserted back into the respective reasoning chains. Completed sequences are moved to the finished set $\mathcal{F}$, while ongoing sequences remain in $\mathcal{S}$ for further processing. By leveraging parallel processing for both generation and refinement steps, the batch inference mechanism enhances system throughput associated with handling multiple inputs concurrently.

\section{Experiments}

\subsection{Tasks and Datasets}
The evaluations used in this experiment include the following two categories:

\textbf{Challenging reasoning tasks:} 
(1) \textbf{GPQA}~\cite{2311_gpqa} is a PhD-level science multiple-choice QA dataset. The questions are authored by domain experts in physics, chemistry, and biology. In our main experiments, we use the highest quality diamond set containing 198 questions, and in Table~\ref{tab:gpqa_extended_set}, we use a more comprehensive extended set containing 546 questions to compare with the performance of human experts.
(2) \textbf{Math benchmarks} include \textbf{MATH500}~\cite{math500}, \textbf{AMC2023}~\footnote{\url{https://huggingface.co/datasets/AI-MO/aimo-validation-amc}}, and \textbf{AIME2024}~\footnote{\url{https://huggingface.co/datasets/AI-MO/aimo-validation-aime}}. MATH500 consists of 500 questions from the MATH test set~\cite{MATH}. AMC2023 and AIME2024 are middle school math competitions covering arithmetic, algebra, geometry, etc., containing 40 and 30 questions respectively. Among these three datasets, MATH500 and AMC are relatively simple, while AIME is more difficult.
(3) \textbf{LiveCodeBench}~\cite{2403_LiveCodeBench} is a benchmark for evaluating LLMs' coding capabilities, consisting of easy, medium, and hard difficulty problems. It collects recently published programming problems from competitive platforms to avoid data contamination. We utilize problems from August to November 2024, comprising 112 problems.

\textbf{Open-domain QA tasks:} 
(1) \textbf{Single-hop QA datasets:} \textbf{Natural Questions (NQ)}~\cite{nq} contains questions from real Google search queries with answers from Wikipedia articles. \textbf{TriviaQA}~\cite{triviaqa} is a large-scale dataset with questions from trivia websites and competitions, featuring complex entity relationships.
(2) \textbf{Multi-hop QA datasets:} \textbf{HotpotQA}~\cite{hotpotqa} is the first large-scale dataset requiring reasoning across multiple Wikipedia paragraphs. \textbf{2WikiMultihopQA (2WIKI)}~\cite{2wiki} provides explicit reasoning paths for multi-hop questions. \textbf{MuSiQue}~\cite{musique} features 2-4 hop questions built from five existing single-hop datasets. \textbf{Bamboogle}~\cite{bamboogle} collects complex questions that Google answers incorrectly to evaluate models' compositional reasoning across various domains.

\subsection{Baselines}
We evaluate our approach against the following baseline methods:

\textbf{Direct Reasoning:} These methods utilize the model's internal knowledge without retrieval. The open-source models include Qwen2.5-32B-Instruct~\cite{qwen2.5}, Qwen2.5-Coder-32B-Instruct~\cite{qwen2.5_coder}, QwQ-32B-Preview~\cite{qwq-32b-preview}, Qwen2.5-72B-Instruct~\cite{qwen2.5}, and Llama3.3-70B-Instruct~\cite{llama3}. Closed-source non-proprietary models include DeepSeek-R1-Lite-Preview~\cite{deepseek-r1}, OpenAI GPT-4o~\cite{gpt_4o_system_card}, and o1-preview~\cite{openai2024openaio1card}. Results for open-source models are based on our implementations, while closed-source model results are sourced from their official releases.

\textbf{Retrieval-augmented Reasoning:} These methods retrieve external information to enhance the reasoning process. We consider two retrieval augmentation approaches: 
\textbf{(1) Standard RAG:} Retrieves the top-10 documents for the original question and inputs them alongside the question into the model for reasoning and answer generation.
\textbf{(2) RAG Agent (RAgent):} Allows the model to decide when to generate queries for retrieval, as detailed in Section~\ref{sec:rag_agent}. To manage the length of retrieved documents, inspired by ReAct~\cite{yao2022react}, we first retrieve the top-10 snippets during reasoning. The model then decides which URLs to obtain for the full documents when necessary.

\subsection{Implementation Details}
For the backbone large reasoning model in Search-o1, we utilize the open-sourced QwQ-32B-Preview~\cite{qwq-32b-preview}. For generation settings, we use a maximum of 32,768 tokens, temperature of 0.7, top\_p of 0.8, top\_k of 20, and a repetition penalty of 1.05 across all models. For retrieval, we employ the Bing Web Search API, setting the region to US-EN and the top-$k$ retrieved documents to 10. We use Jina Reader API to fetch the content of web pages for given URLs. For all retrieval-based methods, following~\cite{2311_gpqa}, we apply a back-off strategy where, when a final answer is not provided, we use the result from direct reasoning. For baseline models not specifically trained for o1-like reasoning, we apply Chain-of-Thought (CoT)~\cite{wei2022chain} prompting to perform reasoning before generating answers. Detailed instructions for all models are provided in Appendix~\ref{app:instructions}. All experiments are conducted on eight NVIDIA A800-80GB GPUs.

\begin{table*}[!t]
\centering
\caption{Main results on challenging reasoning tasks, including PhD-level science QA, math, and code benchmarks. We report Pass@1 metric for all tasks. For models with 32B parameters, the best results are in \textbf{bold} and the second-best are \underline{underlined}. Results from larger or non-proprietary models are in \textcolor{gray!135}{gray} color for reference. Symbol ``$^\dagger$'' indicates results from their official releases.}
\label{tab:reasoning_performance}
\setlength\tabcolsep{1.7pt}
\fontsize{8.1pt}{10.5pt}\selectfont
\begin{tabular}{p{2.65cm}cccccccccccc}
\toprule
\multirow{2}[2]{*}{\textbf{Method}} & \multicolumn{4}{c}{\textbf{GPQA (PhD-Level Science QA)}} & \multicolumn{3}{c}{\textbf{Math Benchmarks}} & \multicolumn{4}{c}{\textbf{LiveCodeBench}} \\
\cmidrule(lr){2-5} \cmidrule(lr){6-8} \cmidrule(lr){9-12}
 & Physics & Chemistry & Biology & Overall & MATH500 & AMC23 & AIME24 & Easy & Medium & Hard & Overall \\
\midrule
\multicolumn{12}{l}{\textit{\textbf{Direct Reasoning (w/o Retrieval)}}} \\
Qwen2.5-32B & 57.0 & 33.3 & 52.6 & 45.5 & 75.8 & 57.5 & 23.3 & 42.3 & 18.9 & \underline{14.3} & 22.3 \\
Qwen2.5-Coder-32B & 37.2 & 25.8 & 57.9 & 33.8 & 71.2 & 67.5 & 20.0 & \underline{61.5} & 16.2 & 12.2 & 25.0 \\
QwQ-32B & 75.6 & 39.8 & 68.4 & 58.1 & 83.2 & \underline{82.5} & \underline{53.3} & \underline{61.5} & \underline{29.7} & \textbf{20.4} & \textbf{33.0} \\
\midrule
Qwen2.5-72B & \textcolor{gray!135}{57.0} & \textcolor{gray!135}{37.6} & \textcolor{gray!135}{68.4} & \textcolor{gray!135}{49.0} & \textcolor{gray!135}{79.4} & \textcolor{gray!135}{67.5} & \textcolor{gray!135}{20.0} & \textcolor{gray!135}{53.8} & \textcolor{gray!135}{29.7} & \textcolor{gray!135}{24.5} & \textcolor{gray!135}{33.0} \\
Llama3.3-70B & \textcolor{gray!135}{54.7} & \textcolor{gray!135}{31.2} & \textcolor{gray!135}{52.6} & \textcolor{gray!135}{43.4} & \textcolor{gray!135}{70.8} & \textcolor{gray!135}{47.5} & \textcolor{gray!135}{36.7} & \textcolor{gray!135}{57.7} & \textcolor{gray!135}{32.4} & \textcolor{gray!135}{24.5} & \textcolor{gray!135}{34.8} \\
DeepSeek-R1-Lite$^\dagger$ & \textcolor{gray!135}{-} & \textcolor{gray!135}{-} & \textcolor{gray!135}{-} & \textcolor{gray!135}{58.5} & \textcolor{gray!135}{91.6} & \textcolor{gray!135}{-} & \textcolor{gray!135}{52.5} & \textcolor{gray!135}{-} & \textcolor{gray!135}{-} & \textcolor{gray!135}{-} & \textcolor{gray!135}{51.6} \\
GPT-4o$^\dagger$ & \textcolor{gray!135}{59.5} & \textcolor{gray!135}{40.2} & \textcolor{gray!135}{61.6} & \textcolor{gray!135}{50.6} & \textcolor{gray!135}{60.3} & \textcolor{gray!135}{-} & \textcolor{gray!135}{9.3} & \textcolor{gray!135}{-} & \textcolor{gray!135}{-} & \textcolor{gray!135}{-} & \textcolor{gray!135}{33.4} \\
o1-preview$^\dagger$ & \textcolor{gray!135}{89.4} & \textcolor{gray!135}{59.9} & \textcolor{gray!135}{65.9} & \textcolor{gray!135}{73.3} & \textcolor{gray!135}{85.5} & \textcolor{gray!135}{-} & \textcolor{gray!135}{44.6} & \textcolor{gray!135}{-} & \textcolor{gray!135}{-} & \textcolor{gray!135}{-} & \textcolor{gray!135}{53.6} \\
\midrule
\multicolumn{12}{l}{\textit{\textbf{Retrieval-augmented Reasoning}}} \\
RAG-Qwen2.5-32B & 57.0 & 37.6 & 52.6 & 47.5 & 82.6 & 72.5 & 30.0 & \underline{61.5} & 24.3 & 8.2 & 25.9 \\
RAG-QwQ-32B & \underline{76.7} & 38.7 & \underline{73.7} & 58.6 & 84.8 & \underline{82.5} & 50.0 & 57.7 & 16.2 & 12.2 & 24.1 \\
RAgent-Qwen2.5-32B & 58.1 & 33.3 & 63.2 & 47.0 & 74.8 & 65.0 & 20.0 & 57.7 & 24.3 & 6.1 & 24.1 \\
RAgent-QwQ-32B & \underline{76.7} & \underline{46.2} & 68.4 & \underline{61.6} & \underline{85.0} & \textbf{85.0} & \textbf{56.7} & \textbf{65.4} & 18.9 & 12.2 & \underline{26.8} \\
\midrule
\multicolumn{12}{l}{\textit{\textbf{Retrieval-augmented Reasoning with Reason-in-Documents}}} \\
\rowcolor[RGB]{236,244,252} 
Search-o1 (Ours) & \textbf{77.9} & \textbf{47.3} & \textbf{78.9} & \textbf{63.6} & \textbf{86.4} & \textbf{85.0} & \textbf{56.7} & 57.7 & \textbf{32.4} & \textbf{20.4} & \textbf{33.0} \\
\bottomrule
\end{tabular}
\end{table*}

\subsection{Results on Challenging Reasoning Tasks}

\paragraph{Main Results.} 
Table~\ref{tab:reasoning_performance} presents Search-o1's performance on complex reasoning tasks, with the main results outlined below:
\begin{enumerate}[leftmargin=1em]
    \item For both settings without retrieval and with retrieval augmentation, \textbf{the large reasoning model QwQ-32B-Preview consistently shows superior performance} compared to traditional instruction-tuned LLMs. The QwQ model with 32B parameters even outperforms larger LLMs such as Qwen2.5-72B and Llama3.3-70B in the direct reasoning setting, demonstrating the effectiveness of the o1-like long CoT approach in complex reasoning.
    \item \textbf{RAgent-QwQ-32B surpasses both standard RAG-based models and direct reasoning QwQ-32B in most tasks}, thanks to its agentic search mechanism, which autonomously retrieves information to supplement the knowledge required for reasoning at each step. Additionally, we find that the non-reasoning model Qwen2.5-32B using agentic RAG performs similarly to standard RAG on GPQA and even shows decreased performance on math and code tasks. This indicates that ordinary LLMs cannot effectively utilize search as a tool to solve complex reasoning tasks.
    \item \textbf{Our Search-o1 further outperforms RAgent-QwQ-32B in most tasks}, demonstrating the effectiveness of our Reason-in-Documents strategy by integrating external knowledge while ensuring that it does not affect the coherence of the original reasoning. Specifically, on average across all five datasets, Search-o1 exceeds RAgent-QwQ-32B and QwQ-32B by 4.7\% and 3.1\%, respectively, and significantly outperforms non-reasoning models Qwen2.5-32B and Llama3.3-70B by 44.7\% and 39.3\%.
\end{enumerate}

\paragraph{Scaling Analysis on Number of Retrieved Documents.}

\begin{wraptable}{r}{0.48\textwidth}
\vspace{-1.5em}
\centering
\caption{Performance comparison with human experts on the GPQA extended set~\cite{2311_gpqa}.}
\vspace{-0.5em}
\label{tab:gpqa_extended_set}
\setlength\tabcolsep{1.7pt} 
\fontsize{8.1pt}{9.8pt}\selectfont
\begin{tabular}{p{2.3cm}cccc}
\toprule
\multirow{2}[2]{*}{\textbf{Method}} & \multicolumn{4}{c}{\textbf{GPQA Extended Set}} \\
\cmidrule(lr){2-5}
& Physics & Chemistry & Biology & Overall \\
\midrule
\multicolumn{5}{l}{\textit{\textbf{Human Experts}}} \\
Physicists         & 57.9 & 31.6 & 42.0 & 39.9 \\
Chemists           & 34.5 & \textbf{72.6} & 45.6 & 48.9 \\
Biologists         & 30.4 & 28.8 & \underline{68.9} & 37.2 \\
\midrule
\multicolumn{5}{l}{\textit{\textbf{Reasoning Models}}} \\
QwQ-32B            & 61.7 & 36.9 & 61.0 & 51.8 \\
RAG-QwQ-32B        & \underline{64.3} & 38.3 & 66.7 & \underline{54.6} \\
\rowcolor[RGB]{236,244,252}
Search-o1 (Ours)   & \textbf{68.7} & \underline{40.7} & \textbf{69.5} & \textbf{57.9} \\

\bottomrule
\end{tabular}
\vspace{-3em}
\end{wraptable}

\begin{figure*}[!t]
\centering
\includegraphics[width=1\linewidth]{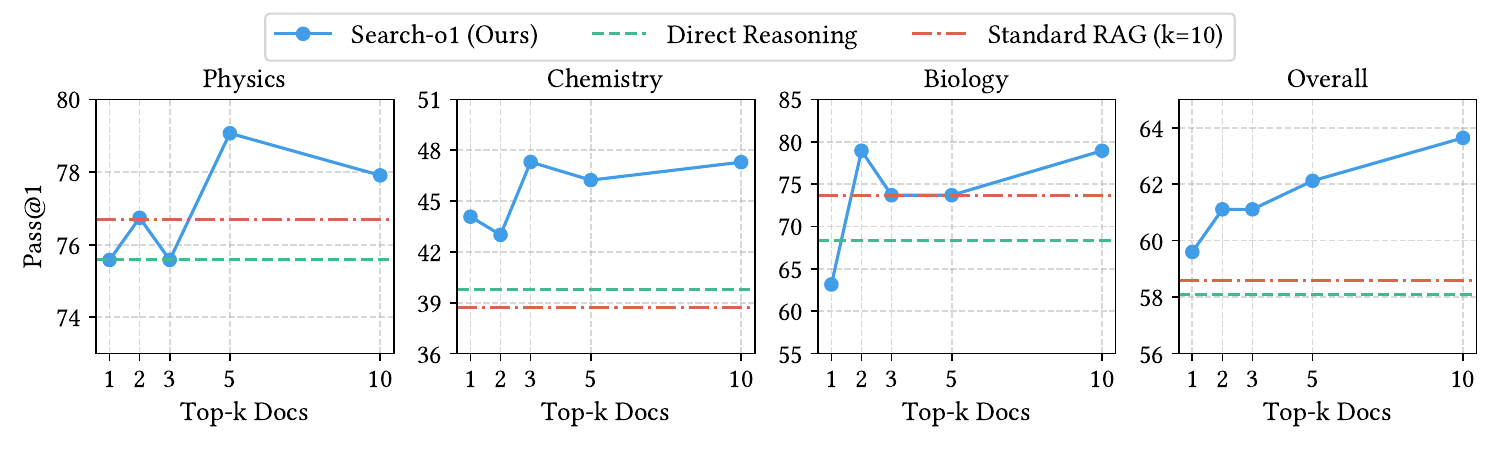}
\caption{
Scaling analysis of top-k retrieved documents utilized in reasoning. All results are based on QwQ-32B-Preview model.
}
\label{fig:gpqa_topk_docs} 
\end{figure*}

In this experiment, we analyze the performance variation with respect to the number of retrieved documents, as shown in Figure~\ref{fig:gpqa_topk_docs}. Our results demonstrate that \textbf{Search-o1 can effectively leverage an increasing number of retrieved documents, leading to improvements in handling complex reasoning tasks}. We also observe that for overall performance, retrieving even one document can surpass Direct Reasoning and standard RAG models that use ten retrieved documents, showcasing the effectiveness of the agentic search and Reason-in-Documents strategies.

\paragraph{Comparison with Human Experts.}
We compare the performance of our Search-o1 with human experts across various domains in the GPQA extended set. Table~\ref{tab:gpqa_extended_set} presents the evaluation of human experts from various disciplines, including physics, chemistry, and biology. Our Search-o1 model outperforms human experts in overall performance (57.9), as well as in both physics (68.7) and biology (69.5), demonstrating superior handling of complex reasoning tasks. While Search-o1 slightly trails chemists in the chemistry subdomain (40.7 vs. 72.6), it still provides a competitive edge overall, particularly in terms of general performance across multiple domains. This highlights the \textbf{effectiveness of Search-o1 in leveraging document retrieval and reasoning to achieve cross-domain performance that rivals or exceeds expert-level capabilities.}

\subsection{Results on Open-Domain QA Tasks}
In addition to the reasoning tasks where LRMs excel, we also explore the performance of our Search-o1 on open-domain QA tasks. Table~\ref{tab:qa_performance} presents the overall results. The key observations are:
\begin{enumerate}[leftmargin=1em]
    \item For direct reasoning without retrieval, the performance of the LRM QwQ-32B is overall similar to the non-reasoning LLM Qwen2.5-32B, with a slight decrease in average EM across all QA datasets (31.3 vs. 30.7). This indicates that \textbf{LRMs do not perform as strongly on open-domain QA tasks as they do on reasoning tasks.}
    \item When employing retrieval-augmented reasoning, retrieval significantly improves performance for both reasoning and non-reasoning models across all tasks, suggesting that \textbf{models have knowledge gaps in these tasks}. Additionally, for the QwQ-32B model, agentic RAG achieves an average EM improvement of 23.2\% over standard RAG on multi-hop QA tasks, demonstrating the \textbf{effectiveness of our agentic RAG strategy in knowledge-based multi-hop QA}. However, we also observe that there is no significant performance change for single-hop tasks (47.8 vs. 47.6 on average EM), as these questions only require information from a single knowledge point without the need for multiple retrievals. \textbf{This also verifies that the agentic search mechanism can better unleash the potential of LRMs in more complex and challenging reasoning tasks.}
    \item \textbf{For our Search-o1, we find that it generally outperforms all baselines on multi-hop tasks.} Specifically, in terms of the average EM metric, our Search-o1 exceeds RAG-QwQ-32B and RAgent-QwQ-32B by 29.6\% and 5.3\%, respectively, demonstrating the effectiveness of our Reason-in-Documents strategy in complex QA tasks. \textbf{This further emphasizes the importance of maintaining consistency between external knowledge and the logical chain of reasoning.}
\end{enumerate}

\begin{table*}[!t]
\centering
\caption{Performance comparison on open-domain QA tasks, including single-hop QA and multi-hop QA datasets. For models with 32B parameters, the best results are in \textbf{bold} and the second-best are \underline{underlined}. Results from larger models are in \textcolor{gray!135}{gray} color for reference.}
\label{tab:qa_performance}
\setlength\tabcolsep{5.75pt}
\fontsize{8.1pt}{10.5pt}\selectfont
\begin{tabular}{p{2.7cm}cccccccccccc}
\toprule
\multirow{3}[3]{*}{\textbf{Method}} & \multicolumn{4}{c}{\textbf{Single-hop QA}} & \multicolumn{8}{c}{\textbf{Multi-hop QA}} \\
\cmidrule(lr){2-5} \cmidrule(lr){6-13}
& \multicolumn{2}{c}{\textbf{NQ}} & \multicolumn{2}{c}{\textbf{TriviaQA}} & \multicolumn{2}{c}{\textbf{HotpotQA}} & \multicolumn{2}{c}{\textbf{2WIKI}} & \multicolumn{2}{c}{\textbf{MuSiQue}} & \multicolumn{2}{c}{\textbf{Bamboogle}} \\
\cmidrule(lr){2-3} \cmidrule(lr){4-5} \cmidrule(lr){6-7} \cmidrule(lr){8-9} \cmidrule(lr){10-11} \cmidrule(lr){12-13}
& EM & F1 & EM & F1 & EM & F1 & EM & F1 & EM & F1 & EM & F1 \\
\midrule
\multicolumn{13}{l}{\textit{\textbf{Direct Reasoning (w/o Retrieval)}}} \\
Qwen2.5-32B & 22.8 & 33.9 & 52.0 & 60.3 & 25.4 & 34.7 & 29.8 & 36.3 & 8.4 & 18.0 & 49.6 & 63.2 \\
QwQ-32B & 23.0 & 33.1 & 53.8 & 60.7 & 25.4 & 33.3 & 34.4 & 40.9 & 9.0 & 18.9 & 38.4 & 53.7 \\
\midrule
Qwen2.5-72B & \textcolor{gray!135}{27.6} & \textcolor{gray!135}{41.2} & \textcolor{gray!135}{56.8} & \textcolor{gray!135}{65.8} & \textcolor{gray!135}{29.2} & \textcolor{gray!135}{38.8} & \textcolor{gray!135}{34.4} & \textcolor{gray!135}{42.7} & \textcolor{gray!135}{11.4} & \textcolor{gray!135}{20.4} & \textcolor{gray!135}{47.2} & \textcolor{gray!135}{61.7} \\
Llama3.3-70B & \textcolor{gray!135}{36.0} & \textcolor{gray!135}{48.7} & \textcolor{gray!135}{68.8} & \textcolor{gray!135}{76.8} & \textcolor{gray!135}{37.8} & \textcolor{gray!135}{49.1} & \textcolor{gray!135}{46.0} & \textcolor{gray!135}{54.2} & \textcolor{gray!135}{14.8} & \textcolor{gray!135}{23.6} & \textcolor{gray!135}{54.4} & \textcolor{gray!135}{67.8} \\
\midrule
\multicolumn{13}{l}{\textit{\textbf{Retrieval-augmented Reasoning}}} \\
RAG-Qwen2.5-32B & 33.4 & \underline{49.3} & \textbf{65.8} & \textbf{79.2} & 38.6 & 50.4 & 31.6 & 40.6 & 10.4 & 19.8 & 52.0 & 66.0 \\
RAG-QwQ-32B & 29.6 & 44.4 & \underline{65.6} & \underline{77.6} & 34.2 & 46.4 & 35.6 & 46.2 & 10.6 & 20.2 & \underline{55.2} & \underline{67.4} \\
RAgent-Qwen2.5-32B & 32.4 & 47.8 & 63.0 & 72.6 & \underline{44.6} & \underline{56.8} & 55.4 & 69.7 & 13.0 & 25.4 & 54.4 & 66.4 \\
RAgent-QwQ-32B & \underline{33.6} & 48.4 & 62.0 & 74.0 & 43.0 & 55.2 & \textbf{58.4} & \underline{71.2} & \underline{13.6} & \underline{25.5} & 52.0 & 64.7 \\
\midrule
\multicolumn{13}{l}{\textit{\textbf{Retrieval-augmented Reasoning with Reason-in-Documents}}} \\
\rowcolor[RGB]{236,244,252}
Search-o1 (Ours) & \textbf{34.0} & \textbf{49.7} & 63.4 & 74.1 & \textbf{45.2} & \textbf{57.3} & \underline{58.0} & \textbf{71.4} & \textbf{16.6} & \textbf{28.2} & \textbf{56.0} & \textbf{67.8} \\
\bottomrule
\end{tabular}
\end{table*}

\section{Conclusion}

In this work, we present Search-o1, a framework that addresses the knowledge insufficiency inherent in large reasoning models (LRMs) by integrating an agentic retrieval-augmented generation mechanism alongside a Reason-in-Documents module. Our approach enables LRMs to autonomously retrieve and seamlessly incorporate external knowledge during the reasoning process, thereby enhancing both the accuracy and coherence of their long-step reasoning capabilities. Comprehensive experiments across diverse complex reasoning tasks in science, mathematics, and coding, as well as multiple open-domain QA benchmarks, demonstrate that Search-o1 consistently outperforms existing retrieval-augmented and direct reasoning methods. Notably, Search-o1 not only surpasses baseline models in handling intricate reasoning challenges but also achieves performance levels comparable to or exceeding human experts in specific domains. These findings underscore the potential of Search-o1 to significantly improve the reliability and versatility of LRMs, paving the way for more trustworthy and effective intelligent systems in complex problem-solving scenarios.

{
\bibliographystyle{plain} 
\bibliography{main}
}

\clearpage
\appendix

\section*{Appendix}


\section{Instruction Templates}
\label{app:instructions}

\subsection{Instructions for Search-o1}
\label{app:instruction_search_o1}

\begin{tcolorbox}[
    colframe = gray,       
    colback = gray!5!white,             
    coltitle = white,                   
    coltext = black,                    
    fonttitle = \bfseries,              
    title = Instruction for Search-o1,  
    boxrule = 1pt,                      
    arc = 2mm,                          
    width = \linewidth,                 
    left = 7pt,                         
    right = 7pt,                        
    top = 5pt,                          
    bottom = 5pt                        
]
\fontsize{8.5pt}{10pt}\selectfont
You are a reasoning assistant with the ability to perform web searches to help you answer the user's question accurately. You have special tools:\\
To perform a search: write {<|begin\_search\_query|>} your query here {<|end\_search\_query|>}.\\
Then, the system will search and analyze relevant web pages, then provide you with helpful information in the format {<|begin\_search\_result|>} ...search results... {<|end\_search\_result|>}.\\
You can repeat the search process multiple times if necessary. The maximum number of search attempts is limited to \{MAX\_SEARCH\_LIMIT\}.\\
Once you have all the information you need, continue your reasoning.\\
Example:\\
Question: ``...''\\
Assistant thinking steps:\\
- I might need to look up details about ...\\
Assistant:\\
{<|begin\_search\_query|>}...{<|end\_search\_query|>}\\
(System returns processed information from relevant web pages)\\
Assistant continues reasoning with the new information...\\
Remember:\\
- Use {<|begin\_search\_query|>} to request a web search and end with {<|end\_search\_query|>}.\\
- When done searching, continue your reasoning.
\end{tcolorbox}

\begin{tcolorbox}[
    colframe = gray,       
    colback = gray!5!white,             
    coltitle = white,                   
    coltext = black,                    
    fonttitle = \bfseries,              
    title = Instruction for Reason-in-Documents,  
    boxrule = 1pt,                      
    arc = 2mm,                          
    width = \linewidth,                 
    left = 7pt,                         
    right = 7pt,                        
    top = 5pt,                          
    bottom = 5pt                        
]
\fontsize{8.5pt}{10pt}\selectfont
Task Instruction:\\
You are tasked with reading and analyzing web pages based on the following inputs: Previous Reasoning Steps, Current Search Query, and Searched Web Pages. Your objective is to extract relevant and helpful information for Current Search Query from the Searched Web Pages and seamlessly integrate this information into the Previous Reasoning Steps to continue reasoning for the original question.\\
Guidelines:\\
1. Analyze the Searched Web Pages:\\
- Carefully review the content of each searched web page.\\
- Identify factual information that is relevant to the Current Search Query and can aid in the reasoning process for the original question.\\
2. Extract Relevant Information:\\
- Select the information from the Searched Web Pages that directly contributes to advancing the Previous Reasoning Steps.\\
- Ensure that the extracted information is accurate and relevant.\\
3. Output Format:\\
- If the web pages provide helpful information for current search query: Present the information beginning with `Final Information` as shown below.\\
Final Information\\
\text{[Helpful information]}\\
- If the web pages do not provide any helpful information for current search query: Output the following text.\\
Final Information\\
No helpful information found.\\
Inputs:\\
- Previous Reasoning Steps:  \\
\{prev\_reasoning\}\\
- Current Search Query:\\
\{search\_query\}\\
- Searched Web Pages:\\
\{document\}\\
Now you should analyze each web page and find helpful information based on the current search query ``\{search\_query\}'' and previous reasoning steps.
\end{tcolorbox}

\subsection{Instructions for Standard RAG}
\label{app:instruction_standard_rag}

\begin{tcolorbox}[
    colframe=gray,       
    colback=gray!5!white,             
    coltitle=white,                   
    coltext=black,                    
    fonttitle=\bfseries,              
    title=Instruction for Standard RAG,  
    boxrule=1pt,                      
    arc=2mm,                          
    width=\linewidth,                 
    left=7pt,                         
    right=7pt,                        
    top=5pt,                          
    bottom=5pt                        
]
\fontsize{8.5pt}{10pt}\selectfont
You are a knowledgeable assistant that utilizes the provided documents to answer the user's question accurately.

Question:  
\{question\}

Documents:  
\{documents\}

Guidelines:

- Analyze the provided documents to extract relevant information. Synthesize the information to formulate a coherent and accurate answer.

- Ensure that your response directly addresses the user's question using the information from the documents.
\end{tcolorbox}

\subsection{Instructions for RAG Agent}
\label{app:instruction_rag_agent}

\begin{tcolorbox}[
    colframe=gray,       
    colback=gray!5!white,             
    coltitle=white,                   
    coltext=black,                    
    fonttitle=\bfseries,              
    title=Instruction for RAG Agent,  
    boxrule=1pt,                      
    arc=2mm,                          
    width=\linewidth,                 
    left=7pt,                         
    right=7pt,                        
    top=5pt,                          
    bottom=5pt                        
]
\fontsize{8.5pt}{10pt}\selectfont
You are a reasoning assistant with the ability to perform web searches and retrieve webpage content to help you answer the user’s question accurately. You have special tools:

- To perform a search: Write `<|begin\_search\_query|>' your query here `<|end\_search\_query|>'.  

  The system will call the web search API with your query and return the search results in the format `<|begin\_search\_result|> ...search results... <|end\_search\_result|>'.  
  
  The search results will include a list of webpages with titles, URLs, and snippets (but not full content).

- To retrieve full page content: After receiving the search results, if you need more detailed information from specific URLs, write `<|begin\_url|> url1, url2, ... <|end\_url|>'.  

  The system will fetch the full page content of those URLs and return it as `<|begin\_full\_page|> ...full page content... <|end\_full\_page|>'.

You can repeat the search process multiple times if necessary. The maximum number of search attempts is limited to \{MAX\_SEARCH\_LIMIT\}.  
You can fetch up to \{MAX\_URL\_FETCH\} URLs for detailed information.

Once you have all the information you need, continue your reasoning.

Example:

Question: ``...''

Assistant thinking steps:
- I need to find out ...

Assistant:
`<|begin\_search\_query|>...<|end\_search\_query|>'

(System returns search results)

Assistant:
`<|begin\_search\_result|> ...search results without full page... <|end\_search\_result|>'

Assistant thinks: The search results mention several URLs. I want full details from one of them.

Assistant:
`<|begin\_url|>http://...<|end\_url|>'

(System returns full page content)

Assistant:
`<|begin\_full\_page|> ...full page content... <|end\_full\_page|>'

Now the assistant has enough information and can continue reasoning.

Remember:

- Use `<|begin\_search\_query|>' to request a web search and end with `<|end\_search\_query|>'.

- Use `<|begin\_url|>' to request full page content and end with `<|end\_url|>'.

- When done retrieving information, continue your reasoning.
\end{tcolorbox}

\subsection{Task-Specific Instructions}
\label{app:instruction_task_specific}

\subsubsection{Open-Domain QA Tasks Instruction}
\label{app:instruction_openqa}

\begin{tcolorbox}[
    colframe=gray,       
    colback=gray!5!white,             
    coltitle=white,                   
    coltext=black,                    
    fonttitle=\bfseries,              
    title=Instruction for Open-Domain QA Tasks,  
    boxrule=1pt,                      
    arc=2mm,                          
    width=\linewidth,                 
    left=7pt,                         
    right=7pt,                        
    top=5pt,                          
    bottom=5pt                        
]
\fontsize{8.5pt}{10pt}\selectfont
Please answer the following question.

You should provide your final answer in the format \textbackslash{}boxed\{YOUR\_ANSWER\}.

Question:  

\{question\}
\end{tcolorbox}

\subsubsection{Math Tasks Instruction}
\label{app:instruction_math}

\begin{tcolorbox}[
    colframe=gray,       
    colback=gray!5!white,             
    coltitle=white,                   
    coltext=black,                    
    fonttitle=\bfseries,              
    title=Instruction for Math Tasks,  
    boxrule=1pt,                      
    arc=2mm,                          
    width=\linewidth,                 
    left=7pt,                         
    right=7pt,                        
    top=5pt,                          
    bottom=5pt                        
]
\fontsize{8.5pt}{10pt}\selectfont
Please answer the following math question.

You should provide your final answer in the format \textbackslash{}boxed\{YOUR\_ANSWER\}.

Question:

\{question\}
\end{tcolorbox}

\subsubsection{Multi-choice Tasks Instruction}
\label{app:instruction_multi_choice}

\begin{tcolorbox}[
    colframe=gray,       
    colback=gray!5!white,             
    coltitle=white,                   
    coltext=black,                    
    fonttitle=\bfseries,              
    title=Instruction for Multi-choice Tasks,  
    boxrule=1pt,                      
    arc=2mm,                          
    width=\linewidth,                 
    left=7pt,                         
    right=7pt,                        
    top=5pt,                          
    bottom=5pt                        
]
\fontsize{8.5pt}{10pt}\selectfont
You are to answer the following multiple-choice question by selecting the correct option.

Your final choice should be one of the letters A, B, C, or D. Do not include any answer content beyond the choice letter.

You should provide your final choice in the format \textbackslash{}boxed\{YOUR\_CHOICE\}.

Question:  
\{question\}
\end{tcolorbox}

\subsubsection{Code Tasks Instruction}
\label{app:instruction_code}

\begin{tcolorbox}[
    colframe=gray,       
    colback=gray!5!white,             
    coltitle=white,                   
    coltext=black,                    
    fonttitle=\bfseries,              
    title=Instruction for Code Tasks,  
    boxrule=1pt,                      
    arc=2mm,                          
    width=\linewidth,                 
    left=7pt,                         
    right=7pt,                        
    top=5pt,                          
    bottom=5pt                        
]
\fontsize{8.5pt}{10pt}\selectfont
Generate a correct Python program that passes all tests for the given problem. You should provide your final code within a Python code block using triple backticks.
\begin{verbatim}
```python
# YOUR CODE HERE
```
\end{verbatim}

Problem Title: \{question\_title\}

Problem Statement:

\{question\}
\end{tcolorbox}

\subsection{Additional Notes}
For all the instructions above, we input them as user prompts, not system prompts. The task-specific instructions in~\ref{app:instruction_task_specific} are used for the QwQ-32B-Preview model. For non-reasoning models like Qwen2.5-32B-Instruct, Qwen2.5-72B-Instruct, and Llama3.3-70B-Instruct, etc., we add a Chain-of-Thought prompt "You should think step by step to solve it." before the question to explicitly make these models reason before giving the final answer.

\section{Case Study}
\label{app:case}
Tables~\ref{tab:case_gpqa}, \ref{tab:case_math}, and \ref{tab:case_qa} present examples of outputs from our Search-o1 model on the GPQA, AMC2023, and HotpotQA datasets, respectively. The model-generated search queries are enclosed within \green{<|begin\_search\_query|>} and \green{<|end\_search\_query|>}, while the refined search results are enclosed within \blue{<|begin\_search\_result|>} and \blue{<|end\_search\_result|>}. We observe that our Reason-in-Documents mechanism provides coherent information that effectively meets the information needs of the model's current reasoning step and seamlessly integrates with the preceding reasoning process.

\begin{table*}[!t]
    \centering
    \caption{An example from Search-o1 on GPQA dataset, with special symbols used in the search queries and search results highlighted in \green{green box} and \blue{blue box}, respectively.}
    \fontsize{9pt}{11pt}\selectfont
    \begin{tabular}{p{0.98\linewidth}}
    \midrule
        \rowcolor{gray!20}\textbf{Example \#1 from GPQA Dataset} \\
    \midrule
        \textbf{Question:} \\
        trans-cinnamaldehyde was treated with methylmagnesium bromide, forming product 1.\\1 was treated with pyridinium chlorochromate, forming product 2.\\3 was treated with (dimethyl(oxo)-l6-sulfaneylidene)methane in DMSO at elevated temperature, forming product 3.\\how many carbon atoms are there in product 3? Choices:\\(A) 10\\(B) 11\\(C) 14\\(D) 12\\
        \textbf{Labeled Answer:} \\
        B \\
    \midrule
        \rowcolor{gray!20}
        \textbf{Model Output by Search-o1}\\
    \midrule
        Alright, so I have this problem here about some chemical reactions, and I need to figure out how many carbon atoms are in the final product, which is product 3. Let's break this down step by step.\\First, there's trans-cinnamaldehyde, and it's being treated with methylmagnesium bromide to form product 1. I know that methylmagnesium bromide is a Grignard reagent, which is nucleophilic and usually adds to carbonyl compounds. So, trans-cinnamaldehyde has a carbonyl group, right? It's an aldehyde, so it has a C=O bond. The Grignard reagent should attack the carbonyl carbon, adding the methyl group and turning the aldehyde into an alcohol, I think. But let me make sure.\\Wait, maybe I should look up what exactly happens when a Grignard reagent reacts with an aldehyde. Let me try to search for that.
        \\\green{<|begin\_search\_query|>}reaction of grignard reagent with aldehyde\green{<|end\_search\_query|>}\\\blue{<|begin\_search\_result|>}Reacting a Grignard reagent with an aldehyde results in the formation of a secondary alcohol, incorporating one additional carbon atom from the Grignard reagent. Therefore, if trans-cinnamaldehyde (9 carbon atoms) reacts with methylmagnesium bromide (1 carbon atom), the resulting product (product 1) will have 10 carbon atoms. Subsequent protonation to form product 2 and further reactions to form product 3 do not change the carbon atom count, assuming no carbon is added or removed in those steps. Hence, product 3 should have 10 carbon atoms.\blue{<|end\_search\_result|>}\\
        Hmm, according to this, product 1 should have 10 carbon atoms, starting from trans-cinnamaldehyde, which has 9 carbon atoms, plus the methyl group from the Grignard reagent, adding one more carbon atom. That makes sense.\\Next, product 1 is treated with pyridinium chlorochromate to form product 2. Pyridinium chlorochromate (PCC) is used for oxidations, particularly for oxidizing primary alcohols to aldehydes or secondary alcohols to ketones. Since product 1 is a secondary alcohol (from the previous step), PCC should oxidize it to a ketone. Again, this shouldn't change the number of carbon atoms; it's just changing the functional group.\\
        ...
        \\Alternatively, perhaps the initial trans-cinnamaldehyde has more or fewer carbon atoms than I thought. Let me confirm its structure.
        \\\green{<|begin\_search\_query|>}structure of trans-cinnamaldehyde\green{<|end\_search\_query|>}\\\blue{<|begin\_search\_result|>}Trans-cinnamaldehyde, also known as (E)-cinnamaldehyde, has the molecular formula C9H8O. It consists of a phenyl ring connected to a propenal group via a double bond, forming a conjugated system that makes the molecule planar. The structure comprises 9 carbon atoms, 8 hydrogen atoms, and 1 oxygen atom, with the functional groups indicating its reactivity in chemical transformations.\blue{<|end\_search\_result|>}\\
        Yes, confirmed: trans-cinnamaldehyde has 9 carbon atoms.\\Now, reacting with methylmagnesium bromide, which provides a methyl group (-CH3), adds one carbon atom, making product 1 have 10 carbon atoms.\\Then, oxidation with PCC to convert the alcohol to a ketone doesn't change the carbon skeleton, so product 2 still has 10 carbon atoms.\\
        ...
        \\Therefore, starting from 9 carbon atoms in trans-cinnamaldehyde, adding one carbon atom from the Grignard reagent to make 10, and then adding another carbon atom in the cyclopropanation to make 11.\\So, the final answer is \boxed{B}\\
    \bottomrule
    \end{tabular}
    \label{tab:case_gpqa}
\end{table*}

\begin{table*}[!t]
    \centering
    \caption{An example from Search-o1 on AMC2023 dataset, with special symbols used in the search queries and search results highlighted in \green{green box} and \blue{blue box}, respectively.}
    \fontsize{9pt}{11pt}\selectfont
    \begin{tabular}{p{0.98\linewidth}}
    \midrule
        \rowcolor{gray!20}\textbf{Example \#2 from AMC2023 Dataset} \\
    \midrule
        \textbf{Question:} \\
        There is a unique sequence of integers $a_1, a_2, \cdots a_{2023}$ such that\[\tan2023x = \frac{a_1 \tan x + a_3 \tan^3 x + a_5 \tan^5 x + \cdots + a_{2023} \tan^{2023} x}{1 + a_2 \tan^2 x + a_4 \tan^4 x \cdots + a_{2022} \tan^{2022} x}\]whenever $\tan 2023x$ is defined. What is $a_{2023}?$\\
        \textbf{Labeled Answer:} \\
        -1 \\
    \midrule
        \rowcolor{gray!20}
        \textbf{Model Output by Search-o1}\\
    \midrule
        Alright, so I've got this problem here involving trigonometric identities, specifically with the tangent function. The equation given is:\\
        ...
        \\
        And I need to find the value of \( a_{2023} \).\\First off, I know that there are multiple-angle identities for tangent, but I'm not sure if there's a direct formula for such a high multiple like 2023. Maybe there's a general formula for \( \tan nx \) in terms of \( \tan x \), and perhaps this problem is asking for the coefficients in that expansion.
        \\
        ...
        \\
        Wait, I remember that tangent of multiple angles can be expressed using the Bernoulli polynomials or something similar, but I'm not sure. Maybe I need to search for a general formula for \( \tan nx \).\\Let me try to search for \"tangent multiple angle formula\" to see if there's a general expression.\\
        \green{<|begin\_search\_query|>}tangent multiple angle formula\green{<|end\_search\_query|>}\\\blue{<|begin\_search\_result|>}The recursive formula for tan(nθ) is given by:\[ \tan(n\theta) = \frac{\tan[(n-1)\theta] + \tan\theta}{1 - \tan[(n-1)\theta]\tan\theta} \]This formula can be used iteratively to compute tan(nθ) for any integer n, making it particularly useful for large values of n, such as n=2023, as encountered in the problem. This approach avoids the complexity of deriving high-degree polynomial expressions directly and leverages the tangent addition formula in a step-by-step manner.\blue{<|end\_search\_result|>}\\
        Okay, so there's a recursive formula for \( \tan(n\theta) \):\[\tan(n\theta) = \frac{\tan[(n-1)\theta] + \tan\theta}{1 - \tan[(n-1)\theta]\tan\theta}\]That seems useful, especially for large \( n \), like 2023. But I'm not sure how to apply this recursively up to 2023 without going through each step, which isn't practical.\\Maybe there's a general expression for \( \tan(nx) \) in terms of \( \tan(x) \) as a rational function.\\
        ...
        \\
        Looking at \( n = 1 \):\[\tan x = \frac{a_1 \tan x}{1} \implies a_1 = 1\]For \( n = 3 \), \( a_3 = -1 \)\\For \( n = 5 \), \( a_5 = 1 \)\\So it seems that \( a_n = (-1)^{(n-1)/2} \) for odd \( n \).\\Wait, let's check:\\For \( n = 1 \), \( (-1)^{(1-1)/2} = (-1)^0 = 1 \), which matches \( a_1 = 1 \)\\For \( n = 3 \), \( (-1)^{(3-1)/2} = (-1)^1 = -1 \), which matches \( a_3 = -1 \)\\For \( n = 5 \), \( (-1)^{(5-1)/2} = (-1)^2 = 1 \), which matches \( a_5 = 1 \)\\So, generalizing, for odd \( n \), \( a_n = (-1)^{(n-1)/2} \)\\Now, 2023 is odd, so \( a_{2023} = (-1)^{(2023-1)/2} = (-1)^{1011} = -1 \)\\Therefore, \( a_{2023} = -1 \)\\I think that's the answer.\\**Final Answer**\[ \boxed{-1} \]\vspace{-0.35cm}\\
    \bottomrule
    \end{tabular}
    \label{tab:case_math}
\end{table*}

\begin{table*}[!t]
    \centering
    \caption{An example from Search-o1 on HotpotQA dataset, with special symbols used in the search queries and search results highlighted in \green{green box} and \blue{blue box}, respectively.}
    \fontsize{9pt}{11pt}\selectfont
    \begin{tabular}{p{0.98\linewidth}}
    \midrule
        \rowcolor{gray!20}\textbf{Example \#3 from HotpotQA Dataset} \\
    \midrule
        \textbf{Question:} \\
        Were Goo Goo Dolls and Echosmith formed in the same city?\\
        \textbf{Labeled Answer:} \\
        no \\
    \midrule
        \rowcolor{gray!20}
        \textbf{Model Output by Search-o1}\\
    \midrule
        Alright, I've got this question: \"Were Goo Goo Dolls and Echosmith formed in the same city?\" Hmm, I don't know much about these bands, so I'll need to look that up.\\First, I should find out where each band was formed. Maybe start with Goo Goo Dolls. I think they're an older band, but I'm not sure. Let me search for their formation place.\\\green{<|begin\_search\_query|>}where was Goo Goo Dolls formed?\green{<|end\_search\_query|>}\\\blue{<|begin\_search\_result|>}The Goo Goo Dolls were formed in Buffalo, New York, in 1986. They initially performed under the name Sex Maggots but changed to Goo Goo Dolls in 1986 for a gig, choosing the name from an ad in True Detective magazine. \blue{<|end\_search\_result|>}\\Okay, so Goo Goo Dolls were formed in Buffalo, New York. Now, I need to find out where Echosmith was formed.\\\green{<|begin\_search\_query|>}where was Echosmith formed?\green{<|end\_search\_query|>}\\\blue{<|begin\_search\_result|>}Echosmith was formed in Chino, California, in February 2009.\blue{<|end\_search\_result|>}\\Chino, California, huh? That sounds different from Buffalo, New York. But maybe I should double-check to make sure I have the right information.\\Let me look up Echosmith again to confirm their formation city.\\\green{<|begin\_search\_query|>}Echosmith formation location\green{<|end\_search\_query|>}\\\blue{<|begin\_search\_result|>}Echosmith was formed in Toluca Lake, Los Angeles, California, in February 2009. The initial confusion about the formation location being Chino, California, or Buffalo, New York, is resolved by multiple recent sources confirming their formation in Los Angeles. Chino might be their hometown, but the band was established in Toluca Lake, Los Angeles.\blue{<|end\_search\_result|>}\\Wait, now it says Toluca Lake, Los Angeles, California. Earlier, I had Chino, California. Maybe there's some confusion here.\\I should check more sources to make sure.\\\green{<|begin\_search\_query|>}Echosmith origin city\green{<|end\_search\_query|>}\\\blue{<|begin\_search\_result|>}Echosmith was born in Chino, California, but formed the band in Toluca Lake, Los Angeles, California, in February 2009. Chino is their hometown, while Toluca Lake is where they officially formed as a band.\blue{<|end\_search\_result|>}\\So, it seems like Echosmith originates from Chino, California, but was formed in Toluca Lake, Los Angeles, California.\\Now, comparing that to Goo Goo Dolls, who were formed in Buffalo, New York.\\Clearly, Buffalo, New York, is different from Toluca Lake, Los Angeles, California.\\But just to be thorough, maybe there's a connection I'm missing.\\Let me see if Echosmith has any ties to Buffalo, New York.\\\green{<|begin\_search\_query|>}Echosmith Buffalo, New York\green{<|end\_search\_query|>}\\\blue{<|begin\_search\_result|>}Echosmith has performed in Buffalo, New York, in the past, as evidenced by a setlist from a concert at Waiting Room on November 1, 2013. However, there are no current tour dates listed for Buffalo in the searched web pages.\blue{<|end\_search\_result|>}\\So, they've played in Buffalo, but that doesn't mean they were formed there.\\Let me see if there's any other connection between the two bands.\\\green{<|begin\_search\_query|>}Goo Goo Dolls and Echosmith connection\green{<|end\_search\_query|>}\blue{<|begin\_search\_result|>}The maximum search limit is exceeded. You are not allowed to search.\blue{<|end\_search\_result|>}\\Hmm, it seems I've reached my search limit. Based on the information I have so far, Goo Goo Dolls were formed in Buffalo, New York, and Echosmith was formed in Toluca Lake, Los Angeles, California. These are different cities, so the answer should be no, they were not formed in the same city.\\Therefore, I can confidently say that Goo Goo Dolls and Echosmith were not formed in the same city.\\**Final Answer**\[ \boxed{\text{no}} \]\vspace{-0.35cm}\\
    \bottomrule
    \end{tabular}
    \label{tab:case_qa}
\end{table*}

\end{CJK}
\end{document}